\documentclass[preprint]{elsarticle}
    
    \usepackage{amssymb}
    \usepackage{amsmath}
    \usepackage{graphicx}
    \usepackage{microtype}
    \usepackage{cleveref}
    \usepackage{booktabs}
    \usepackage{multirow}
    \usepackage{makecell}
    
    \journal{Pattern Recognition}
    
\begin{document}
    
    \begin{frontmatter}
    
\title{Handwriting Trajectory Recovery via Autoregressive Ordered Stroke Instance Prediction}

\author[inst1,inst2]{En-Guang Wang}

\author[inst1]{Yan-Ming Zhang\corref{cor1}}
\ead{ymzhang@nlpr.ia.ac.cn}

\author[inst1]{Fei Yin}

\author[inst1,inst2,inst3]{Cheng-Lin Liu}

\cortext[cor1]{Corresponding author.}
\affiliation[inst1]{
organization={Institute of Automation, Chinese Academy of Sciences},
city={Beijing},
postcode={100190},
country={China}
}

\affiliation[inst2]{
organization={School of Information Science and Technology, ShanghaiTech University},
city={Shanghai},
postcode={201210},
country={China}
}

\affiliation[inst3]{
organization={School of Artificial Intelligence, University of Chinese Academy of Sciences},
city={Beijing},
postcode={100049},
country={China}
}
    
\begin{abstract}
Handwriting trajectory recovery aims to infer the dynamic writing
process hidden behind a static handwritten image.
Since offline handwriting preserves only the final spatial ink pattern,
temporal information such as stroke order, writing direction, and
pen-tip motion is lost, making recovery inherently ambiguous. Existing
learning-based methods often directly predict the complete character
trajectory without explicitly exploiting the stroke-level organization
of handwriting. We argue that recovering the writing process should
follow the writing process itself. Accordingly, we propose a
two-stage framework that first recovers ordered stroke instances and
then reconstructs continuous within-stroke motion. The first stage
integrates stroke extraction and stroke-order recovery through
autoregressive ordered stroke prediction, while direction-related
structural cues further support within-stroke trajectory generation.
Experiments on Chinese handwriting show that the proposed ordered
prediction is more effective than post-hoc stroke ordering. Even without trajectory simplification, our full-point model achieves
numerically better results than those reported by all compared
baselines,
while a controlled analysis shows that trajectory sampling density
substantially affects measured recovery performance. Additional experiments demonstrate generalization
to unseen Chinese character categories and cross-language extensibility
to English and Tamil handwriting.
\end{abstract}

    
    \end{frontmatter}

\section{Introduction}
\label{sec:introduction}

\begin{figure*}[t]
    \centering
    \setlength{\tabcolsep}{2.5pt}
    \renewcommand{\arraystretch}{0.8}
    \begin{tabular}{ccccc}
        \includegraphics[width=0.185\textwidth]{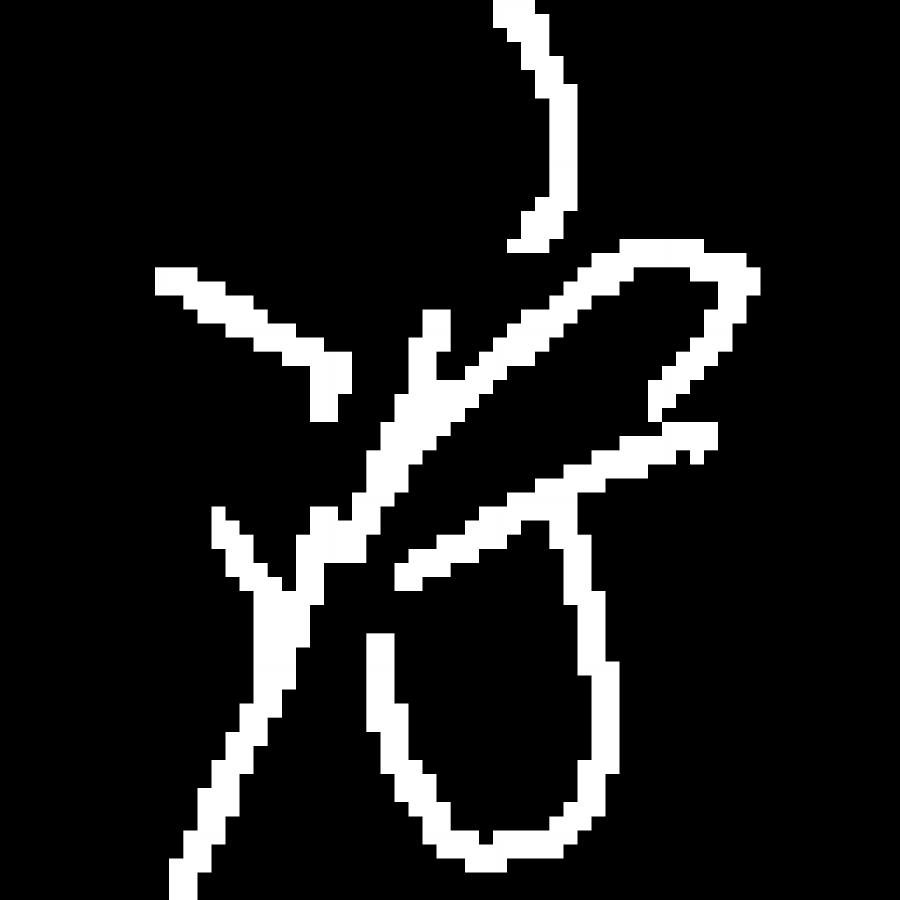} &
        \includegraphics[width=0.185\textwidth]{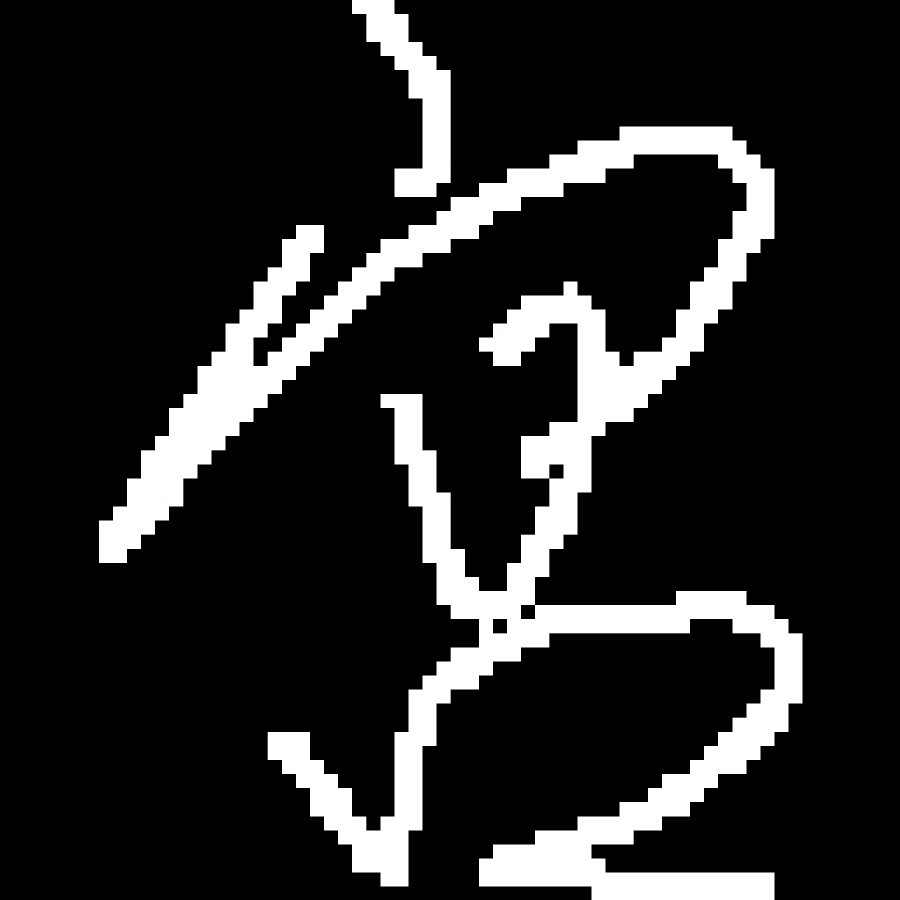} &
        \includegraphics[width=0.185\textwidth]{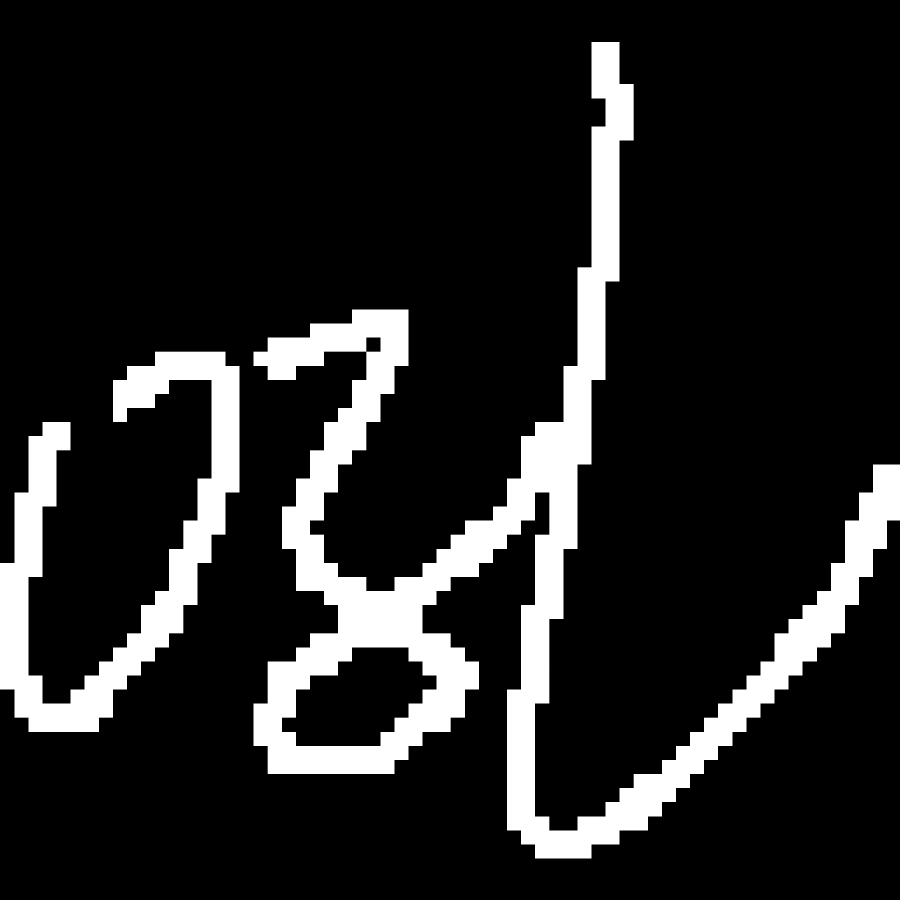} &
        \includegraphics[width=0.185\textwidth]{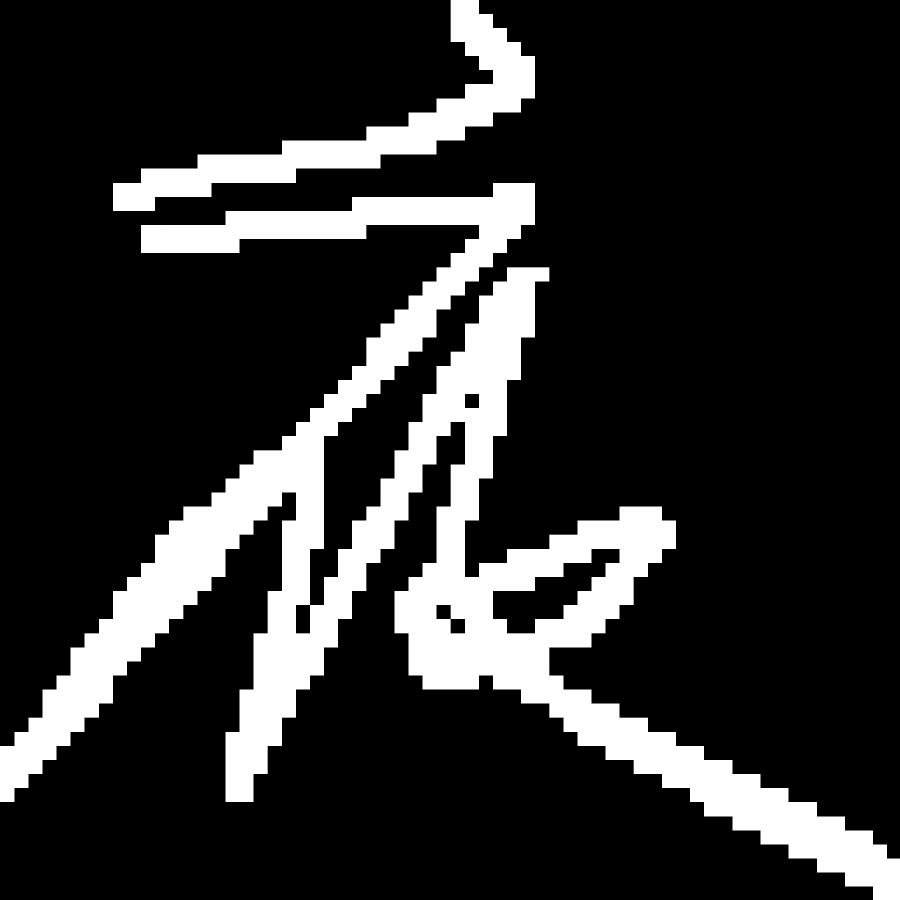} &
        \includegraphics[width=0.185\textwidth]{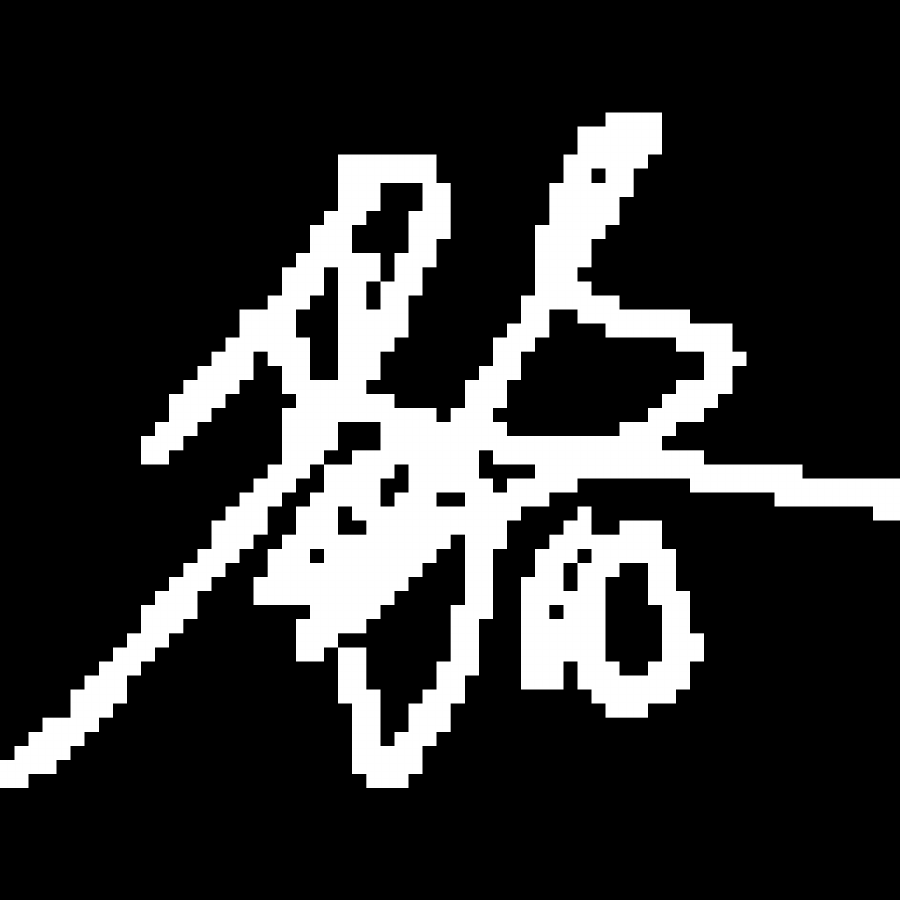}
        \\[-2pt]
        \includegraphics[width=0.185\textwidth]{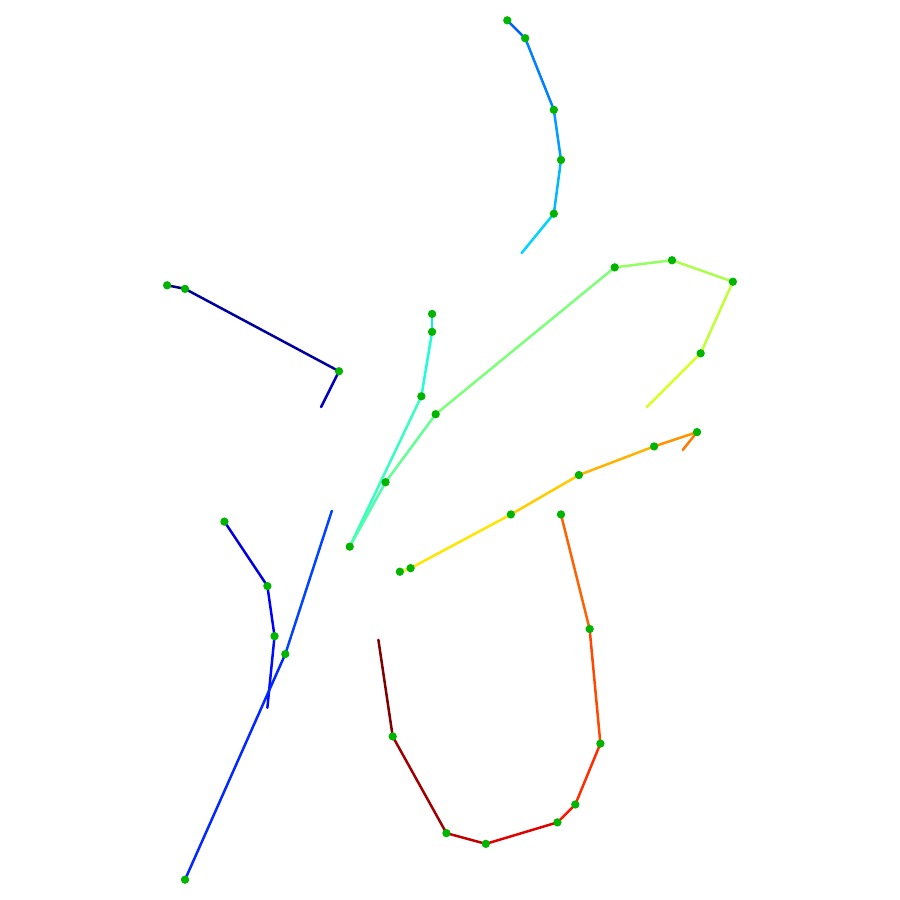} &
        \includegraphics[width=0.185\textwidth]{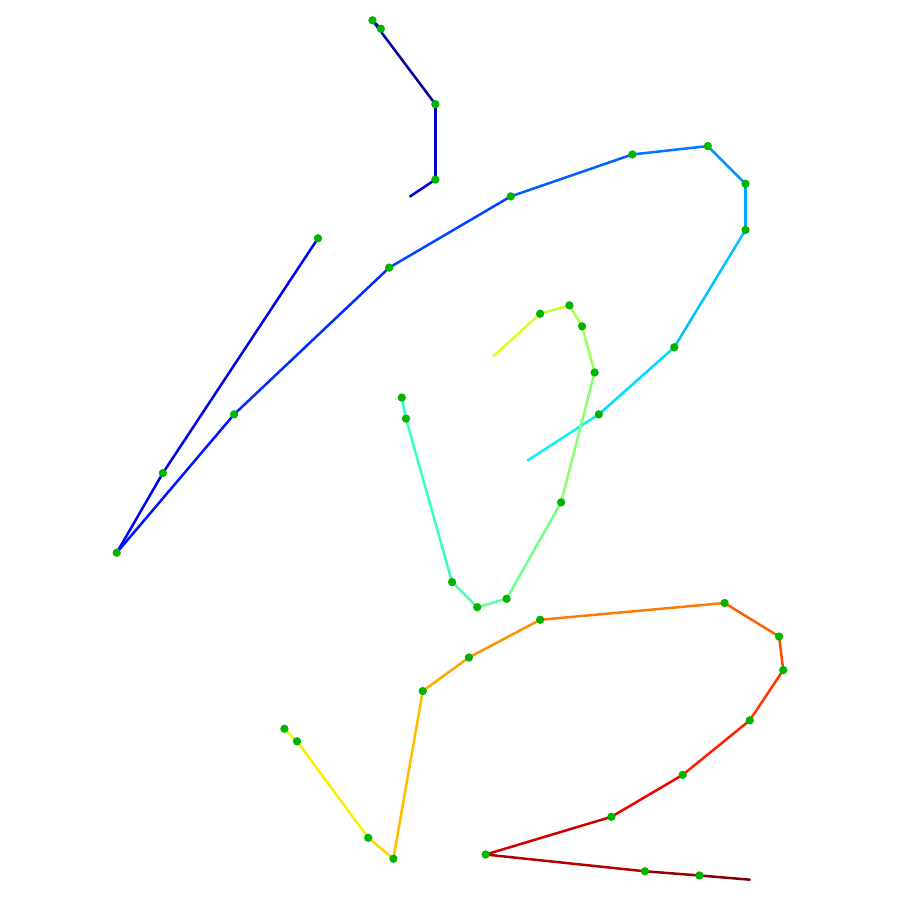} &
        \includegraphics[width=0.185\textwidth]{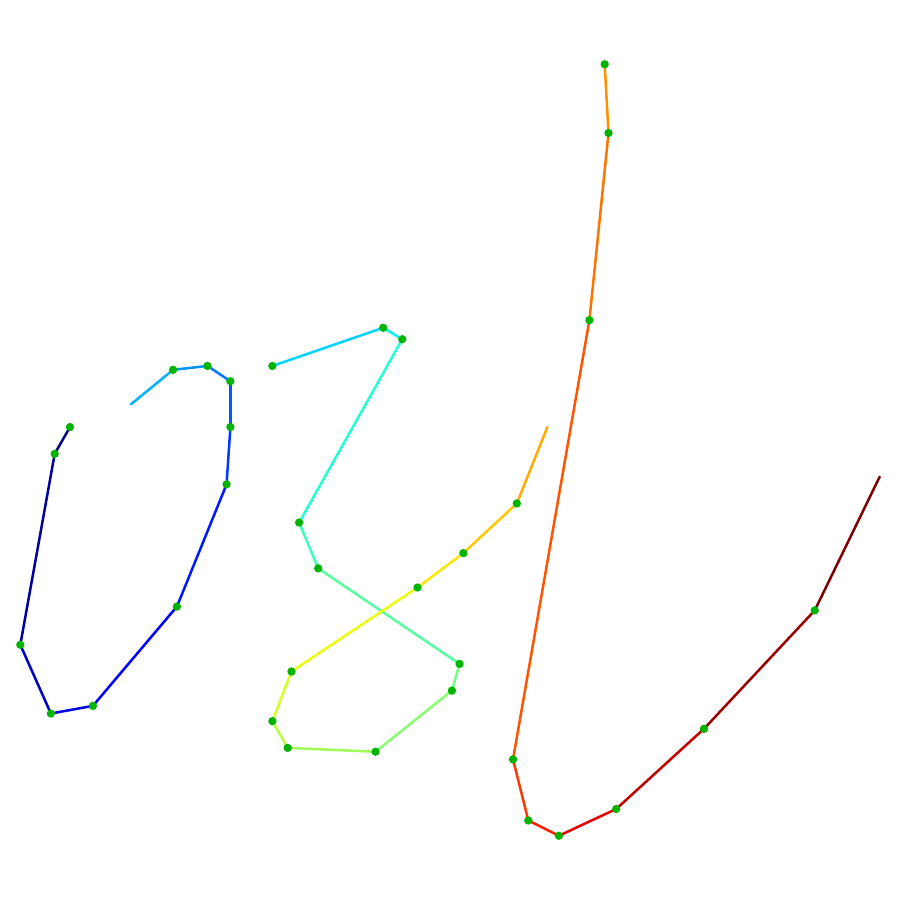} &
        \includegraphics[width=0.185\textwidth]{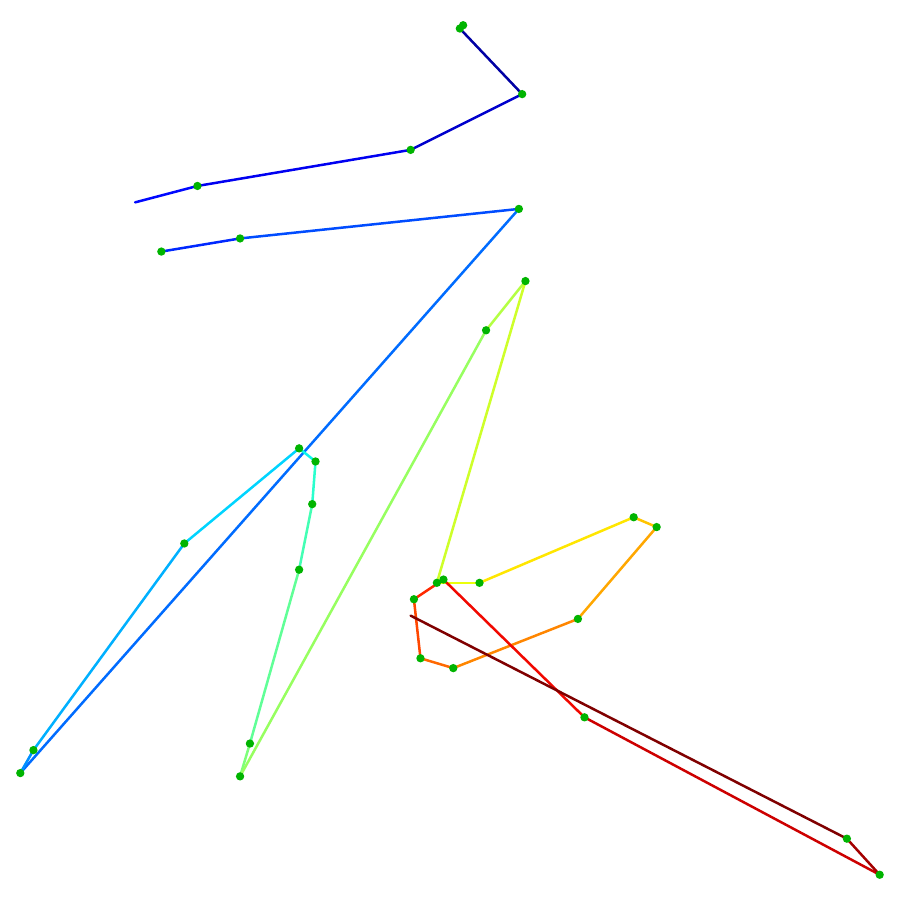} &
        \includegraphics[width=0.185\textwidth]{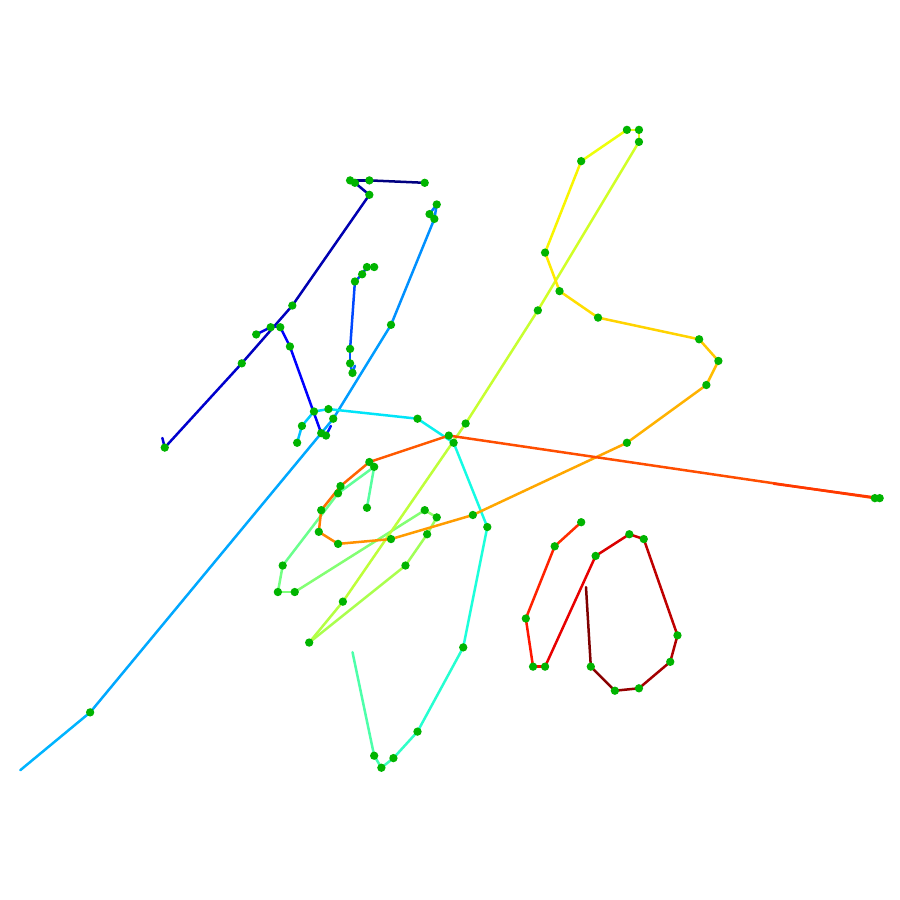}
    \end{tabular}
    \caption{
        Examples of offline handwritten character images (top) and
        their corresponding online writing trajectories (bottom).
        Green dots denote trajectory sampling points, and the trajectory
        color progresses from cool to warm tones as writing time
        advances.
    }
    \label{fig:writing_process}
\end{figure*}

Handwriting is inherently a dynamic process generated by continuous
pen-tip movements, whereas an offline handwritten image preserves only
the final spatial ink pattern produced by this process
~\cite{plamondon2000survey}. Unlike online handwriting, which records
pen-tip coordinates and their temporal order, offline handwriting loses
dynamic information such as stroke order, writing direction, and
continuous pen-tip motion during image formation
~\cite{plamondon2000survey,kato2000drawing}. Offline-to-online
handwriting trajectory recovery therefore aims not merely to
reconstruct the observed character shape, but to infer the underlying
writing process that produced the static image. As illustrated in
Fig.~\ref{fig:writing_process}, the offline image shows only the final
ink pattern, whereas the corresponding online trajectory explicitly
reveals the pen-tip motion and its temporal evolution.

Inferring this dynamic process from a static image is inherently
ambiguous~\cite{sumi2019crossvae,nguyen2020kanji}. Different stroke
orders, starting positions, and writing directions may produce similar
or even nearly identical final glyphs
~\cite{kato2000drawing,qiao2006strokeorder}. The ambiguity becomes
more severe for multi-stroke characters containing intersections,
contacts, overlaps, or cursive connections, where local ink appearance
may be insufficient to determine stroke ownership and temporal
precedence~\cite{wang2021icme,wang2021phd}. Consequently, the central
challenge is not simply to find a geometric path compatible with the
observed image, but to recover the stroke structure, inter-stroke
order, and continuous within-stroke pen-tip motion that together
constitute the writing process.

Early studies primarily relied on explicit structural reasoning.
Skeleton analysis and global optimization were used to recover
plausible writing paths from static ink patterns
~\cite{jaeger1996traces,kato2000drawing}, while graph search and hierarchical character structures were further
employed for stroke-order recovery in multi-stroke Chinese characters
~\cite{qiao2006strokeorder,cao2008sequence}.
Deep learning later
shifted the problem toward direct trajectory prediction. DED-Net
formulated trajectory recovery as end-to-end image-to-sequence
generation~\cite{bhunia2018dednet}, whereas Cross-VAE modeled the
conversion between offline and online handwriting modalities
~\cite{sumi2019crossvae}. Subsequent studies employed attention- or
Transformer-based architectures for variable-length and complex
trajectory recovery
~\cite{nguyen2020kanji,lin2024trajectorytransformer}. In parallel,
stroke decomposition and ordering explicitly introduced stroke-level
structure~\cite{wang2021icme}, while stroke segmentation improved
stroke parsing from static handwriting images
~\cite{gong2024strokeseg,mo2024yolo}.

Despite these advances, a common gap remains. Direct whole-character
generation typically treats the complete trajectory as the primary
prediction target, leaving stroke instance parsing, inter-stroke
ordering, and long-sequence trajectory generation to be resolved
implicitly within a character-level decoding process. In contrast,
methods that explicitly introduce stroke structure often follow an
\emph{extract-then-order} paradigm, in which stroke elements are first
determined and only afterward arranged into a writing sequence
~\cite{wang2021icme}. The former does not fully exploit the intrinsic
hierarchy of handwriting, whereas the latter treats stroke order as a
post-hoc operation that cannot participate in stroke-instance
formation. This raises the central question addressed in this work:
\emph{how can the lost dynamic information be progressively recovered
according to the organization of the actual writing process?}

We argue that recovering the writing process should follow the writing
process itself rather than directly predicting the final trajectory.
Handwriting has a natural hierarchy: continuous pen-tip movements form
individual strokes, and temporally ordered strokes further compose a
complete character~\cite{cao2008sequence,wang2021phd}. Accordingly,
we first recover ordered stroke instances and then recover continuous
pen-tip motion within each stroke. Rather than producing an unordered
stroke set followed by post-hoc ordering, the first stage conditions
instance prediction on the preceding writing state so that each
decoding step directly forms the next stroke, integrating stroke
extraction and stroke-order recovery within the same prediction
process.

To further recover within-stroke motion, the first stage jointly
predicts stroke start and end points as direction-aware auxiliary
structural cues, with the predicted start point transferred across
stages to determine the writing origin of the current stroke. The
second stage then reconstructs subsequent pen-tip motion from this
initialized writing origin. Finally, the recovered stroke trajectories
are concatenated according to their predicted order to form the
complete character trajectory.

Extensive experiments validate both the design choices and the overall
effectiveness of the proposed framework. On Chinese handwriting, the
autoregressive ordered formulation achieves better stroke-order
recovery and downstream trajectory recovery than post-hoc ordering
pipelines, while the full-point model achieves numerically better
results than those reported by all compared baselines across DTW,
LDTW, AIoU, and LPIPS. The framework further generalizes to unseen
Chinese character categories and extends to English and Tamil
handwriting, while controlled experiments reveal the substantial
influence of trajectory sampling density on measured recovery
performance.

The main contributions of this work are summarized as follows:
\begin{itemize}

    \item \textbf{Writing-process-oriented trajectory recovery.}
    We propose a two-stage framework that follows the writing
    process itself, decomposing complete trajectory recovery into
    ordered stroke instance prediction and within-stroke trajectory
    generation.

    \item \textbf{Autoregressive ordered stroke instance prediction.}
    We reformulate query-based unordered set prediction as
    autoregressive ordered stroke prediction, jointly modeling stroke
    extraction and stroke-order recovery during instance formation
    rather than through post-hoc ordering.

    \item \textbf{Direction-aware supervision and start-point initialization.}
    We introduce stroke start/end points as auxiliary structural
    supervision for ordered stroke prediction and transfer the predicted
    start point across stages to initialize within-stroke trajectory
    generation.

    \item \textbf{Full-point trajectory recovery and sampling-density analysis.}
    We evaluate complete trajectory recovery while retaining all original
    trajectory samples and systematically analyze the influence of
    trajectory sampling density; even without trajectory simplification,
    our method achieves numerically better results than those reported by
    all compared baselines.

\end{itemize}

\section{Related Work}
\label{sec:related_work}

\subsection{Offline-to-Online Handwriting Trajectory Recovery}
\label{subsec:related_trajectory_recovery}

Offline-to-online handwriting trajectory recovery has evolved from
explicit structure-driven reconstruction to learning-based sequence
generation. Early approaches treated writing recovery as structural
inference from static handwriting
~\cite{plamondon2000survey}. Jaeger~\cite{jaeger1996traces}
constructed a line graph from a handwriting skeleton and recovered the
writing path through global optimization, while Kato and
Yasuhara~\cite{kato2000drawing} combined graph traversal with heuristic
constraints for single-stroke drawing-order recovery. Related work on
graph matching~\cite{bunke1997graph} and document symbol
representation~\cite{cordella2000symbol} provided general structural
tools for reasoning about spatial configurations.

For multi-stroke Chinese characters, recovery further requires
reasoning about stroke direction and inter-stroke order. Qiao et
al.~\cite{qiao2006strokeorder} recovered stroke order through graph
search under character-structure constraints. Cao et
al.~\cite{cao2008sequence} further introduced a hierarchical
character representation and recovered writing sequences by estimating
stroke directions and precedence relations after structural
decomposition. These methods made stroke structure and order explicit,
but relied heavily on skeletonization, handcrafted constraints, and
staged inference.

Deep learning subsequently shifted the problem toward data-driven
trajectory prediction. PTMP~\cite{zhao2018ptmp} iteratively predicted
the next pen-tip position, whereas DED-Net~\cite{bhunia2018dednet}
formulated trajectory recovery as end-to-end image-to-sequence
generation. Cross-VAE~\cite{sumi2019crossvae} learned conversion
between offline and online handwriting through a shared latent space,
while Nguyen et al.~\cite{nguyen2020kanji} employed an
attention-based encoder--decoder for variable-length Kanji trajectory
recovery. PEN-Net~\cite{chen2022pennet} further addressed complex
handwriting trajectory recovery, and Trajectory Transformer
~\cite{lin2024trajectorytransformer} modeled contextual dependencies
in long trajectory sequences. More recent approaches include
FINet~\cite{zhu2025finet}, which combines spatial encoding with
temporal trajectory decoding, and InkSight
~\cite{mitrevski2025inksight}, which explores vision-language priors
for offline-to-online handwriting conversion.

Although learning-based methods substantially reduce dependence on
handcrafted structural rules, many formulations still predict the
complete character trajectory as the primary output sequence. As a result, stroke instance recovery, inter-stroke ordering, and
within-stroke pen-tip motion remain largely entangled within the
overall character-level generation process. This
motivates recovering the writing process through explicit stroke-level
organization before generating the complete trajectory.

\subsection{Stroke Decomposition and Ordering for Trajectory Recovery}
\label{subsec:related_stroke_ordering}

For multi-stroke handwriting, recovering the spatial stroke structure
does not by itself determine the writing process. After decomposition,
the resulting strokes or stroke segments form an unordered collection,
whereas handwriting is generated in a specific temporal order
~\cite{cao2008sequence}. Wang's stroke-analysis study
~\cite{wang2021phd} further highlights the importance of stroke
structure and ordering in handwritten Chinese character analysis.
Consequently, converting decomposed stroke elements into a writing
sequence constitutes an explicit ordering problem.

General set-to-sequence studies provide a useful modeling perspective.
Order Matters~\cite{vinyals2016ordermatters} showed that mappings from
unordered sets to order-sensitive outputs require dedicated sequence
modeling. Pointer Networks~\cite{vinyals2015pointer} introduced an
autoregressive attention mechanism that outputs indices of input
elements, naturally supporting variable-length permutation problems.

For handwriting trajectory recovery, Wang et
al.~\cite{wang2021icme} applied learned ordering prediction to
pre-extracted stroke segments. Their framework estimated ordering
relations using a convolutional Pointer Network and subsequently
recovered the writing order through heuristic search. This work
demonstrated the importance of explicitly modeling stroke order, but
also represents the typical extract-then-order paradigm: stroke
elements are determined before ordering is performed.

Such a separation limits the interaction between stroke parsing and
stroke-order reasoning. Errors introduced during stroke extraction
propagate directly to the subsequent ordering stage, while writing
history cannot participate in forming the stroke instances themselves.
This motivates modeling stroke identity and stroke order within the
same prediction process rather than treating ordering only as a
post-hoc operation.

\subsection{Stroke Segmentation and Instance-Level Stroke Parsing}
\label{subsec:related_stroke_parsing}

Stroke decomposition provides the spatial units required for
structure-aware trajectory recovery. However, intersections, contacts,
and overlapping regions make stroke parsing difficult because multiple
strokes may share similar local appearance. Stroke-Seg
~\cite{gong2024strokeseg} addresses such ambiguity through
character-specific priors and multi-label segmentation, although
semantic-level outputs do not inherently distinguish repeated stroke
instances.

Instance segmentation explicitly represents individual strokes.
CCSE~\cite{liu2022ccse} formulated Chinese character stroke extraction
with a two-stage framework based on Mask R-CNN
~\cite{he2017maskrcnn}. More recently, Mo and
Wei~\cite{mo2024yolo} employed a YOLO-based segmentation framework
with coordinate-aware attention and enhanced multi-scale feature
fusion for fine-grained Chinese stroke segmentation. These studies
show that instance-level modeling is useful when individual stroke
identities must be preserved.

Related segmentation research has also explored auxiliary geometric
cues. Joint semantic segmentation and boundary detection
~\cite{zhen2020boundary} improves spatial representation through
boundary supervision, while joint contour-point and semantic
prediction~\cite{zhang2020contour} introduces explicit geometric
points into instance segmentation. Such cues enrich spatial structure,
but do not directly represent the writing direction of a handwriting
stroke.

Beyond handwriting-specific stroke segmentation, DETR
~\cite{carion2020detr} introduced query-based set prediction for
generic object detection. Mask2Former~\cite{cheng2022mask2former}
extended query-based mask classification to general image
segmentation, while MaskDINO~\cite{li2023maskdino} unified
query-based detection and segmentation with instance-mask prediction.
These methods were not developed for handwriting stroke extraction;
rather, they provide generic query-based formulations in which each
query represents an instance-level prediction. Their standard
set-prediction formulation, however, is permutation-invariant and does
not encode the temporal writing order among the predicted instances.

Therefore, existing stroke segmentation methods mainly recover spatial
stroke structure, while conventional query-based instance prediction
also remains unordered with respect to the writing process. This gap
motivates our autoregressive ordered stroke instance prediction, in
which the preceding writing state conditions the formation of the next
stroke so that stroke parsing and stroke-order recovery are integrated
within the same prediction process.

\section{Method}
\label{sec:method}

\subsection{Overview}
\label{sec:method_overview}

An offline handwritten image preserves only the final spatial ink
pattern, while our goal is to recover the dynamic writing process
behind this static observation. We argue that \emph{recovering the
writing process should follow the writing process itself rather than
directly generating the final character trajectory}. In actual
handwriting, strokes are produced sequentially over time, and each
stroke is further formed by continuous pen-tip motion. The writing
process therefore exhibits a natural hierarchy between inter-stroke
temporal organization and within-stroke motion.

Given an offline character image \(I\), the first stage recovers an
ordered stroke-instance sequence
\[
\mathcal{Y}
=
(y_1,y_2,\ldots,y_N),
\]
where \(N\) denotes the number of strokes and the \(t\)-th stroke
instance is represented as
\[
y_t
=
\left(
M_t,
B_t,
c_t,
\mathbf{u}_t,
\mathbf{v}_t
\right).
\]
Here, \(M_t\), \(B_t\), and \(c_t\) denote the stroke mask, bounding
box, and stroke-validity confidence, respectively, while
\(\mathbf{u}_t\) and \(\mathbf{v}_t\) denote the stroke start and end
points.

\begin{figure*}[t]
    \centering
    \includegraphics[
        width=0.98\textwidth,
        trim=10pt 10pt 10pt 10pt,
        clip
    ]{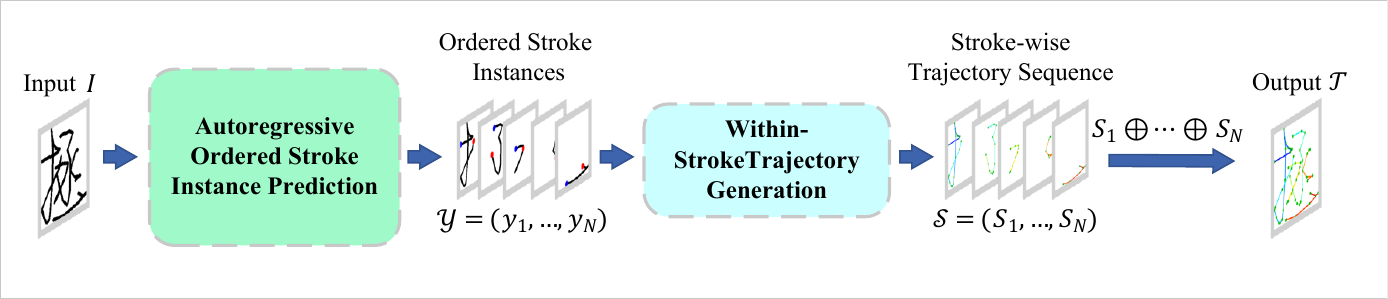}
    \caption{
        Overview of the proposed two-stage offline-to-online
        handwriting trajectory recovery framework. The offline image
        \(I\) is first converted into an ordered stroke-instance
        sequence \(\mathcal{Y}\), followed by stroke-wise trajectory
        recovery \(\mathcal{S}\) and final trajectory concatenation
        \(\mathcal{T}\).
    }
    \label{fig:overall_framework}
\end{figure*}

The second stage further recovers the continuous pen-tip motion within
each ordered stroke. The trajectory of the \(t\)-th stroke is
represented as
\[
S_t
=
\left(
\mathbf{p}_{t,1},
\mathbf{p}_{t,2},
\ldots,
\mathbf{p}_{t,L_t}
\right),
\]
where \(\mathbf{p}_{t,l}=(x_{t,l},y_{t,l})\) denotes the \(l\)-th
trajectory point and \(L_t\) is the trajectory length. Collectively,
the recovered stroke trajectories constitute the ordered stroke-wise
trajectory sequence \(\mathcal{S}\). The complete character trajectory
is then obtained by concatenating them according to the recovered
writing order:
\[
\mathcal{T}
=
\bigoplus_{t=1}^{N} S_t,
\]
where \(\oplus\) denotes trajectory concatenation in stroke order.
Accordingly, as illustrated in Fig.~\ref{fig:overall_framework}, the
overall recovery process can be summarized as
\[
I
\longrightarrow
\mathcal{Y}
\longrightarrow
\mathcal{S}
\longrightarrow
\mathcal{T}.
\]

The first stage recovers stroke structure and inter-stroke temporal
organization through autoregressive ordered stroke instance prediction,
whereas the second stage reconstructs continuous within-stroke motion.
The two stages are trained independently and are described in detail
below. For notation simplicity, recovered quantities are written
without a hat; ground-truth annotations are distinguished using the
superscript \((\cdot)^*\).

\subsection{Autoregressive Ordered Stroke Instance Prediction}
\label{sec:stage1}

Standard query-based instance segmentation formulates target instances as an unordered set and therefore does not explicitly represent the writing
order among stroke instances. Directly applying this formulation to
trajectory recovery typically requires all strokes to be extracted
first and then ordered by an additional model. Consequently, stroke-order reasoning can only be applied to already
formed stroke instances and cannot participate in instance formation
itself.

In contrast, actual handwriting progresses stroke by stroke according
to the current writing state. We therefore reformulate query-based
unordered set prediction as \emph{autoregressive ordered stroke
instance prediction}, allowing the preceding writing state to directly
condition the formation of the next stroke. The ordered prediction
process is factorized as
\[
p(\mathcal{Y}\mid I)
=
\prod_{t=1}^{N}
p
\left(
y_t
\mid
I,y_{<t}
\right),
\]
where
$
y_{<t}
=
(y_1,\ldots,y_{t-1})
$
denotes the writing history recovered before the \(t\)-th decoding
step. Therefore, the model does not select the next stroke from an
already formed candidate set. Instead, it directly forms \(y_t\)
conditioned on the complete character image and the preceding writing
state, allowing stroke parsing and stroke-order recovery to be jointly
modeled during instance formation.
\begin{figure*}[t]
    \centering
    \includegraphics[
        width=0.98\textwidth,
        trim=10pt 10pt 10pt 10pt,
        clip
    ]{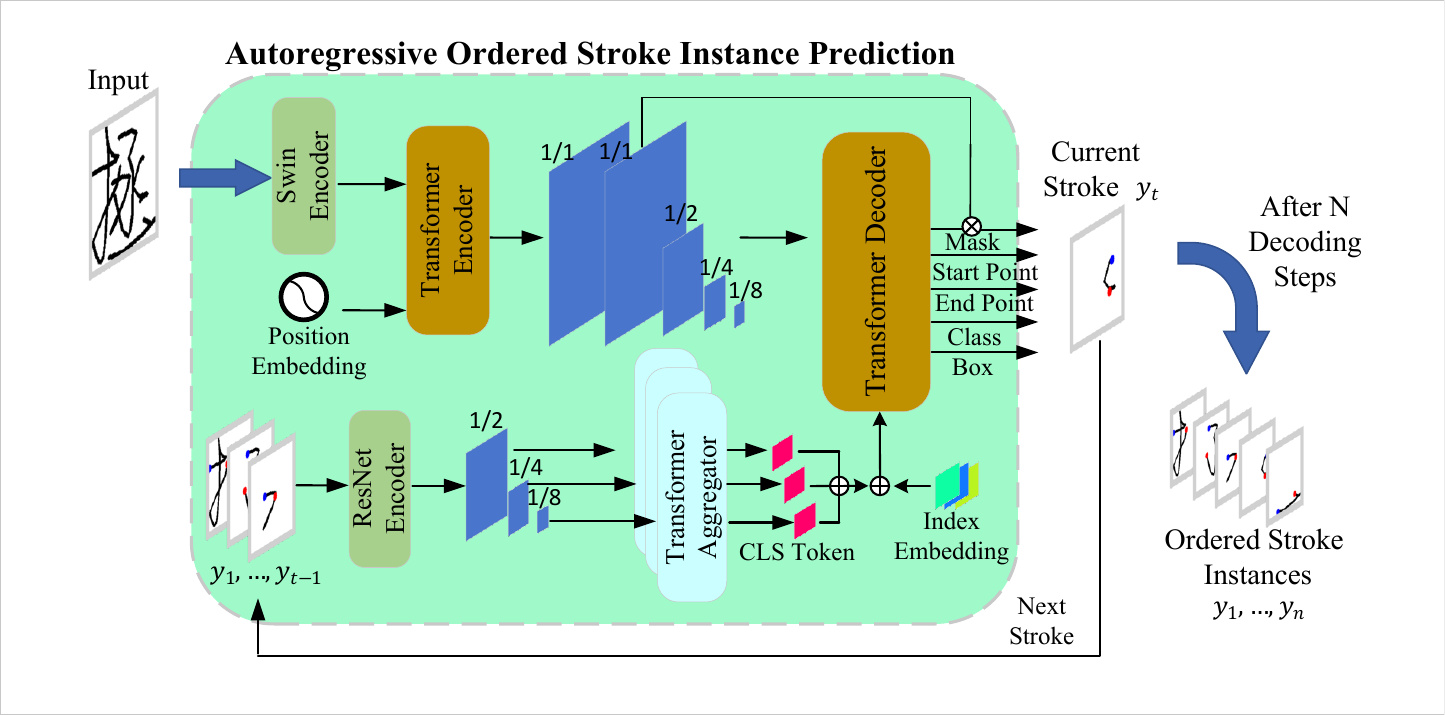}
    \caption{
        Architecture of autoregressive ordered stroke instance
        prediction. Previously recovered strokes form the writing
        history for predicting the current stroke \(y_t\), which is
        then appended to the history for subsequent decoding.
    }
    \label{fig:stage1}
\end{figure*}

\noindent\textbf{Writing-history encoding.}
As shown in Fig.~\ref{fig:stage1}, the strokes recovered before the
\(t\)-th decoding step constitute the current writing history.
Ground-truth historical strokes are used during training, whereas
previously predicted strokes are used during inference.

For the \(i\)-th historical stroke, its mask is first processed by a
shared ResNet encoder~\cite{he2016resnet} to obtain three-scale spatial
features:
\[
F_i^{(k)}
=
E_{\mathrm{str}}^{(k)}
(M_i),
\qquad
k=1,2,3.
\]
At each scale, a Transformer aggregator uses a learnable
\texttt{[CLS]} token to aggregate the spatial feature into a
stroke-level representation:
\[
h_i^{(k)}
=
\operatorname{Agg}_k
\left(
F_i^{(k)},
\mathbf{c}_{\mathrm{cls}}^{(k)}
\right),
\qquad
k=1,2,3.
\]
The three representations are subsequently fused:
\[
z_i
=
\operatorname{Fuse}
\left(
h_i^{(1)},
h_i^{(2)},
h_i^{(3)}
\right).
\]
Since stroke appearance alone does not identify its temporal position
in the writing history, a stroke-index embedding is further added:
\[
\widetilde{z}_i
=
z_i+\rho_i,
\]
where \(\rho_i\) denotes the embedding of the \(i\)-th writing
position. Thus, \(\widetilde{z}_i\) encodes both the spatial structure
of the historical stroke and its position in the preceding writing
sequence.

\noindent\textbf{History-conditioned next-stroke decoding.}
In parallel, the complete character image is processed by a Swin
Encoder~\cite{liu2021swin} and a Transformer
Encoder~\cite{vaswani2017attention} to obtain multi-scale character
features:
\[
\mathcal{F}_I^{\mathrm{enc}}
=
E_{\mathrm{img}}(I).
\]
At autoregressive step \(t\), the history representations form the
query sequence
\[
Q_t^{(0)}
=
\left[
\widetilde{z}_0,
\widetilde{z}_1,
\ldots,
\widetilde{z}_{t-1}
\right],
\]
where \(\widetilde{z}_0\) denotes the initial writing-state token.
The Transformer Decoder interacts with the complete character features:
\[
\bar{Q}_t
=
D_{\mathrm{dec}}
\left(
Q_t^{(0)},
\mathcal{F}_I^{\mathrm{enc}}
\right),
\qquad
q_t
=
[\bar{Q}_t]_{\mathrm{last}}.
\]
The history representations describe what has already been recovered,
while the complete-character features provide the remaining spatial
evidence in the static image. Their interaction enables the decoder to
actively retrieve information relevant to the next stroke and form the
current stroke representation \(q_t\).

\noindent\textbf{Stroke-instance and structural-point prediction.}
The current stroke representation is mapped to task-specific outputs:
\[
\begin{alignedat}{3}
c_t
&=
\sigma\!\left(f_{\mathrm{cls}}(q_t)\right),
\qquad&
B_t
&=
f_{\mathrm{box}}(q_t),
\qquad&
e_t
&=
f_{\mathrm{mask}}(q_t),
\\
M_t
&=
\sigma\!\left(e_t^{\top}F_I^{\mathrm{mask}}\right),
\qquad&
\mathbf{u}_t
&=
f_{\mathrm{start}}(q_t),
\qquad&
\mathbf{v}_t
&=
f_{\mathrm{end}}(q_t).
\end{alignedat}
\]
Here, \(e_t\) is the mask embedding projected from the current stroke
representation, and \(F_I^{\mathrm{mask}}\) is the pixel-level mask
feature produced by the pixel decoder. Their interaction produces the
mask \(M_t\) of the current stroke.

A static stroke mask specifies the spatial region occupied by a stroke
but does not explicitly encode its writing direction. We therefore use
the first and last points of the ground-truth online trajectory as
direction-related structural supervision. For the ground-truth
trajectory
$S_t^*
=
\left(
\mathbf{p}_{t,1}^*,
\ldots,
\mathbf{p}_{t,L_t}^*
\right)$,
the start- and end-point targets are
\[
\mathbf{u}_t^*
=
\mathbf{p}_{t,1}^*,
\qquad
\mathbf{v}_t^*
=
\mathbf{p}_{t,L_t}^*.
\]
Both points provide auxiliary structural supervision for the first
stage. The predicted start point \(\mathbf{u}_t\) is further
transferred to the second stage to initialize within-stroke trajectory
generation, whereas the predicted end point \(\mathbf{v}_t\) is used
only for first-stage supervision.

During inference, each valid predicted stroke is appended to the
writing history and conditions the subsequent decoding step. The
autoregressive process terminates when the stroke-validity confidence
falls below a predefined threshold.

\subsection{Start-Point-Initialized Within-Stroke Trajectory Generation}
\label{sec:stage2}

The stroke masks recovered by the first stage determine the spatial
support of individual strokes but do not determine which endpoint
serves as the writing origin. Starting from opposite endpoints of the
same static stroke may produce spatially plausible trajectories with
opposite writing directions. This ambiguity is particularly important
in the second stage because it observes an isolated stroke without the
complete-character structure and preceding writing state available in
the first stage.

We therefore transfer the start point \(\mathbf{u}_t\), inferred by
the first stage from the complete character and writing history, to
initialize within-stroke trajectory generation. During
complete-character inference, the first trajectory point of stroke
\(t\) is directly initialized as
\[
\mathbf{p}_{t,1}
=
\mathbf{u}_t.
\]
The second stage consequently focuses on recovering the subsequent
continuous pen-tip motion from an explicitly determined writing origin.

\begin{figure*}[t]
    \centering
    \includegraphics[
        width=0.98\textwidth,
        trim=10pt 10pt 10pt 10pt,
        clip
    ]{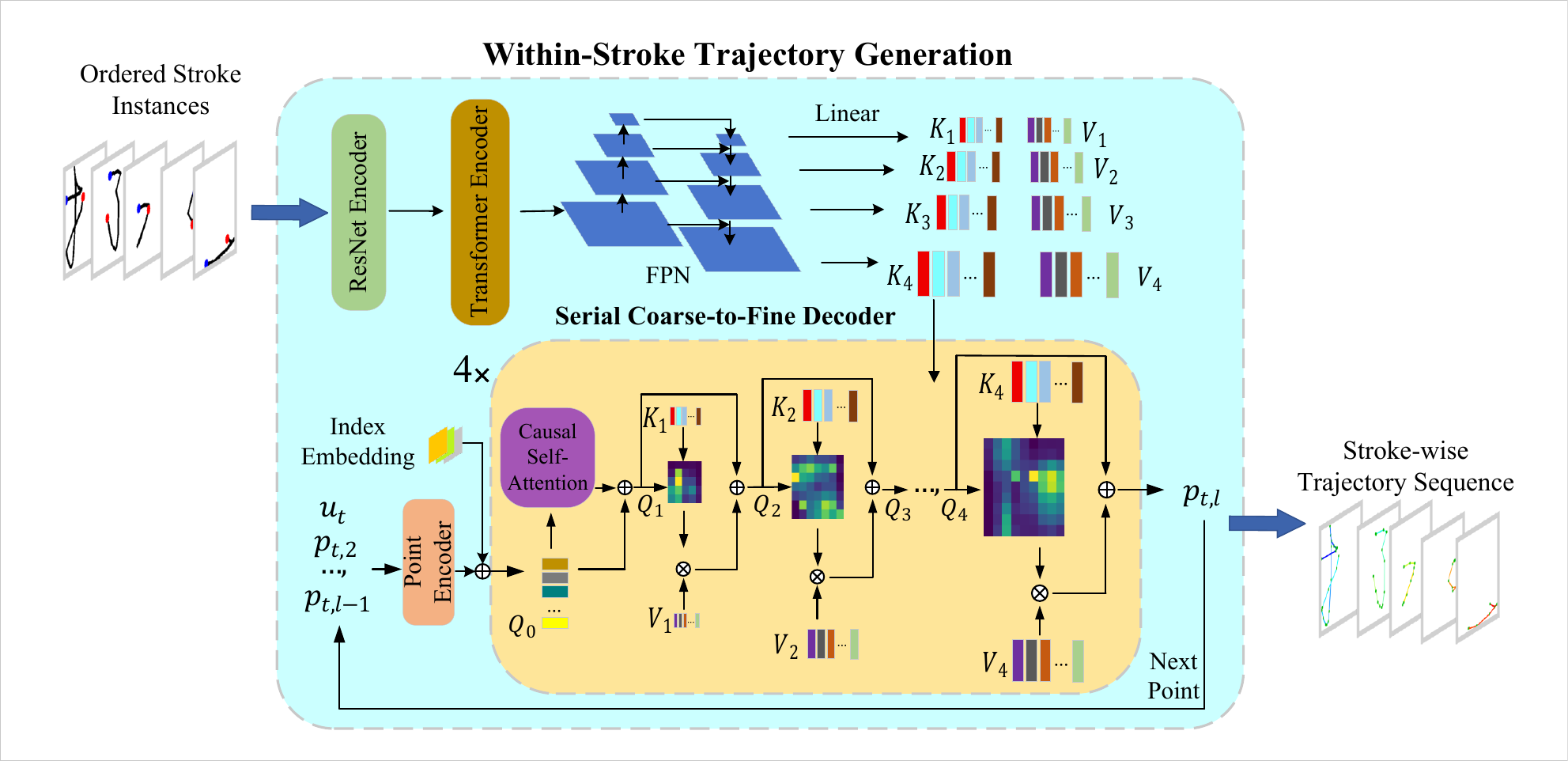}
    \caption{
        Architecture of the start-point-initialized within-stroke
        trajectory generation model with serial coarse-to-fine
        multi-scale decoding.
    }
    \label{fig:stage2}
\end{figure*}

\noindent\textbf{Stroke-level multi-scale visual encoding.}
As shown in Fig.~\ref{fig:stage2}, the \(t\)-th ordered stroke instance
is processed by a ResNet Encoder, a Transformer Encoder, and an
FPN~\cite{lin2017fpn} to construct four stroke-level visual features
from coarse to fine:
\[
\mathcal{F}_t^{\mathrm{st}}
=
\left\{
F_t^{P_5},
F_t^{P_4},
F_t^{P_3},
F_t^{P_2}
\right\}.
\]
Coarser features provide a larger effective receptive field for
capturing the overall stroke geometry and continuation tendency,
whereas higher-resolution features provide progressively finer
information about local bends, boundaries, and spatial locations.

For subsequent cross-attention, the four feature levels are projected
into corresponding key--value pairs:
\[
\begin{aligned}
(K_t^{(1)},V_t^{(1)})
&=
\phi_1(F_t^{P_5}),
\qquad&
(K_t^{(2)},V_t^{(2)})
&=
\phi_2(F_t^{P_4}),
\\
(K_t^{(3)},V_t^{(3)})
&=
\phi_3(F_t^{P_3}),
\qquad&
(K_t^{(4)},V_t^{(4)})
&=
\phi_4(F_t^{P_2}).
\end{aligned}
\]

\noindent\textbf{Start-point-conditioned trajectory-history encoding.}
The second stage is trained independently using ground-truth strokes
and trajectories with teacher forcing. During complete-character
inference, the start point predicted by the first stage is used as
\(\mathbf{p}_{t,1}\).

To predict the \(l\)-th trajectory point, the available trajectory
prefix is embedded as
\[
X_{t,l}
=
\left[
\mathbf{b}_{\mathrm{BOS}}+\pi_0,\,
E_{\mathrm{pt}}(\mathbf{p}_{t,1})+\pi_1,\,
\ldots,\,
E_{\mathrm{pt}}(\mathbf{p}_{t,l-1})+\pi_{l-1}
\right],
\]
where \(\mathbf{b}_{\mathrm{BOS}}\) is a learnable sequence-start
token, \(E_{\mathrm{pt}}(\cdot)\) denotes the Point Encoder, and
\(\pi_l\) denotes the trajectory-index embedding. During training,
ground-truth trajectory prefixes are used in the same formulation.

Causal self-attention then models the temporal dependencies within the
available trajectory prefix:
\[
Q_{t,l}^{(0)}
=
\operatorname{CausalSelfAttn}
\left(
X_{t,l}
\right).
\]
The causal mask prevents the decoder from accessing future trajectory
points.

\noindent\textbf{Serial coarse-to-fine trajectory decoding.}
Rather than fusing all visual scales in parallel, the trajectory query
interacts with the four visual feature levels sequentially from coarse
to fine. The update at scale \(m\) is written as
\[
Q_{t,l}^{(m)}
=
Q_{t,l}^{(m-1)}
+
\operatorname{CrossAttn}
\left(
Q_{t,l}^{(m-1)},
K_t^{(m)},
V_t^{(m)}
\right),
\qquad
m=1,2,3,4.
\]
Here, \(m=1,2,3,4\) correspond to \(P_5,P_4,P_3,P_2\), respectively.
The first cross-attention therefore retrieves coarse structural
information from \(P_5\); its updated trajectory representation is
then passed to the next interaction with \(P_4\), followed by \(P_3\)
and finally \(P_2\).

This serial decoding progressively accumulates multi-scale visual
evidence within the same trajectory representation. Coarse-scale
features first capture the global stroke geometry and continuation
tendency, whereas finer features subsequently refine local bends and
precise spatial locations. Since the representation updated at one
scale directly serves as the query for the next scale, local point
prediction remains conditioned on the global stroke structure captured
earlier. For clarity, standard Transformer operations such as
normalization and feed-forward layers are omitted from the above
formulation. The \(4\times\) notation in Fig.~\ref{fig:stage2}
indicates that the coarse-to-fine decoder block is stacked for four
layers.

\noindent\textbf{Trajectory-point prediction and termination.}
After trajectory-history modeling and coarse-to-fine visual
interaction, the representation for predicting the \(l\)-th point is
obtained from the final decoding state:
\[
r_{t,l}
=
\left[
Q_{t,l}^{(4)}
\right]_{\mathrm{last}}.
\]
The trajectory coordinate and point validity are then predicted as
\[
\mathbf{p}_{t,l}
=
\sigma
\left(
f_{\mathrm{coord}}(r_{t,l})
\right),
\qquad
\gamma_{t,l}
=
\sigma
\left(
f_{\mathrm{valid}}(r_{t,l})
\right).
\]
Generation of the current stroke terminates when \(\gamma_{t,l}\)
falls below a predefined threshold, yielding
\[
S_t
=
\left(
\mathbf{p}_{t,1},
\ldots,
\mathbf{p}_{t,L_t}
\right).
\]

\subsection{Optimization Objectives}
\label{sec:loss}

The two stages are optimized independently. The first-stage objective
jointly supervises stroke validity, instance masks, bounding boxes, and
the start/end structural points:
\[
\begin{aligned}
\mathcal{L}_{\mathrm{stage1}}
={}&
\lambda_{\mathrm{cls}}\mathcal{L}_{\mathrm{cls}}
+
\lambda_{\mathrm{mask}}\mathcal{L}_{\mathrm{mask}}
+
\lambda_{\mathrm{dice}}\mathcal{L}_{\mathrm{dice}}
+
\lambda_{\mathrm{box}}\mathcal{L}_{\mathrm{box}}
\\
&+
\lambda_{\mathrm{giou}}\mathcal{L}_{\mathrm{giou}}
+
\lambda_{\mathrm{s}}\mathcal{L}_{\mathrm{start}}
+
\lambda_{\mathrm{e}}\mathcal{L}_{\mathrm{end}}.
\end{aligned}
\]
Here, \(\mathcal{L}_{\mathrm{cls}}\) supervises stroke validity;
\(\mathcal{L}_{\mathrm{mask}}\) and
\(\mathcal{L}_{\mathrm{dice}}\) supervise instance masks;
\(\mathcal{L}_{\mathrm{box}}\) and
\(\mathcal{L}_{\mathrm{giou}}\) supervise bounding boxes; and
\(\mathcal{L}_{\mathrm{start}}\) and
\(\mathcal{L}_{\mathrm{end}}\) are \(L_1\) losses for the stroke start
and end points. Intermediate Transformer Decoder layers also receive
auxiliary supervision.

The second-stage objective jointly supervises trajectory coordinates
and trajectory termination:
\[
\mathcal{L}_{\mathrm{stage2}}
=
\mathcal{L}_{\mathrm{coord}}
+
\lambda_{\mathrm{valid}}
\mathcal{L}_{\mathrm{valid}}.
\]
Here, \(\mathcal{L}_{\mathrm{coord}}\) is an \(L_1\) loss over valid
ground-truth trajectory points, and
\(\mathcal{L}_{\mathrm{valid}}\) is a binary cross-entropy loss for
point validity. An additional invalid position is
appended after the last ground-truth point to learn trajectory termination; this
position receives validity supervision only and no coordinate supervision.

\section{Experiments}
\label{sec:experiments}
\subsection{Experimental Setup}
\label{subsec:experimental_setup}

\noindent\textbf{Datasets.}
The main experiments are conducted on
CASIA-OLHWDB1.1~\cite{liu2009casia}. Offline character images are
rendered from the original online trajectories. The first stage uses
ordered stroke-instance annotations, including stroke masks, bounding
boxes, stroke order, and start/end points, whereas the second stage
uses individual stroke images and their corresponding within-stroke
trajectory sequences.

For unseen-category evaluation, we construct a category-disjoint
Chinese test set from CASIA-OLHWDB1.2 containing 3,025 character
categories not present in the CASIA-OLHWDB1.1 training set, resulting
in 7,900 test samples. We additionally evaluate English handwriting
from CASIA-OLHWDB following PEN-Net~\cite{chen2022pennet} and
Trajectory Transformer~\cite{lin2024trajectorytransformer}. The
English character categories are included in training, while the test
writers are different from the training writers. For Tamil, we use the
IWFHR 2006 Online Tamil Handwritten Character Recognition Competition
dataset~\cite{hp2006tamil}. Following the settings of PEN-Net and
Trajectory Transformer, both stages are fine-tuned on Tamil training
data before evaluation.

\medskip

\noindent\textbf{Trajectory sampling settings.}
For the main Chinese and English experiments, all trajectory samples
in the original online annotations are retained during both training
and evaluation without simplification. We refer to this configuration
as the \emph{full-point trajectory setting}. Tamil experiments follow
the sampling and preprocessing settings adopted in PEN-Net and
Trajectory Transformer.

To examine the influence of trajectory sampling density, we further
construct RDP-simplified Chinese data using the
Ramer--Douglas--Peucker algorithm
~\cite{ramer1972polygonal,douglas1973reduction}. For the
complete-pipeline experiment with RDP $=3$, the offline images and
online trajectory targets are constructed consistently from the same
simplified trajectories. We further isolate the effect of sampling
density in Section~\ref{subsec:point_simplification} using
ground-truth strokes and character-level trajectory evaluation.

\medskip

\noindent\textbf{Evaluation metrics.}
We evaluate the recovered writing process at three levels. For stroke
instance prediction, Mask AP and Mask $AP_{50}$ are reported following
the COCO protocol~\cite{lin2014coco}; Mask AP is averaged over IoU
thresholds from 0.50 to 0.95 at intervals of 0.05, while Mask
$AP_{50}$ uses an IoU threshold of 0.50.

Stroke-order recovery is evaluated using strict character-level order
accuracy (Order Acc.). A character is counted as correct only when the
predicted and ground-truth stroke numbers are equal, all strokes are
matched one-to-one at mask IoU $\geq 0.50$, and the complete matched
sequence exactly follows the ground-truth writing order. Any missed,
redundant, split, merged, or incorrectly ordered stroke therefore makes
the character incorrect.

For complete trajectory recovery, we use DTW
~\cite{sakoe1978dtw}, LDTW~\cite{chen2022pennet}, AIoU
~\cite{chen2022pennet}, and LPIPS~\cite{zhang2018lpips}. DTW and LDTW
measure trajectory alignment, while AIoU and LPIPS evaluate spatial
coverage and perceptual glyph similarity, respectively. Higher values
are better for Mask AP, Mask $AP_{50}$, Order Acc., and AIoU, whereas
lower values are better for DTW, LDTW, and LPIPS.

\medskip

\noindent\textbf{Implementation details.}
Experiments are conducted on four NVIDIA RTX A6000 GPUs using
Python 3.8.2, PyTorch 1.9.0, and CUDA 11.1. Unless otherwise specified,
both stages use Adam with an initial learning rate of
$1\times10^{-4}$ and cosine annealing. For the main CASIA experiments,
the first and second stages are trained for 500,000 and 991,000
iterations with batch sizes of 64 and 512, respectively.

For the first stage, the loss weights are set to
$\lambda_{\mathrm{cls}}=4$,
$\lambda_{\mathrm{mask}}=\lambda_{\mathrm{dice}}
=\lambda_{\mathrm{box}}=\lambda_{\mathrm{s}}
=\lambda_{\mathrm{e}}=5$, and
$\lambda_{\mathrm{giou}}=2$.
For the second stage, the coordinate loss uses unit weight and
$\lambda_{\mathrm{valid}}=0.2$.
The first-stage Transformer decoder contains nine layers, while the
second-stage serial coarse-to-fine decoder contains four layers.
During inference, both the stroke-validity and point-validity
thresholds are set to $0.5$.

Images are rendered at $64\times64$ resolution. Line widths of 1, 2,
and 3 pixels are randomly sampled during training, while a fixed width
of 2 pixels is used for CASIA validation and testing.

\subsection{Stroke Extraction and Stroke Order Recovery}
\label{subsec:stroke_extraction_order}

We first investigate whether stroke order should be modeled during
stroke-instance prediction or imposed afterward on independently
extracted stroke instances. To construct conventional
extract-then-order pipelines, YOLO26~\cite{jocher2026yolo26},
Mask2Former~\cite{cheng2022mask2former}, and MaskDINO
~\cite{li2023maskdino} are used to extract unordered stroke instances,
which are subsequently processed by the same Transformer Pointer
ordering model. In contrast, our method directly generates stroke
instances in writing order by conditioning each prediction step on the
preceding writing state. For reference, Conv-Ptr-Net
~\cite{wang2021icme} and Transformer Pointer are additionally evaluated
using ground-truth strokes as oracle inputs to isolate their ordering
capability from stroke-extraction errors.

\begin{table*}[t] \centering \caption{ Comparison of stroke extraction, stroke-order recovery, and downstream complete trajectory recovery. Predicted-stroke post-hoc pipelines use the same Transformer Pointer ordering model and the same within-stroke trajectory generator. For the post-hoc pipelines, the writing origin is predicted by the second-stage generator from each isolated stroke. } \label{tab:stroke_extraction_order} \small \setlength{\tabcolsep}{4pt} \renewcommand{\arraystretch}{1.12} \resizebox{0.98\textwidth}{!}{ \begin{tabular}{@{}lcccccc@{}} \toprule \multirow{2}{*}{Pipeline} & \multicolumn{2}{c}{Stroke Extraction / Order Recovery} & \multicolumn{4}{c}{Complete Trajectory Recovery} \\ \cmidrule(lr){2-3} \cmidrule(lr){4-7} & Mask $AP_{50}$ $\uparrow$ & Order Acc. (\%) $\uparrow$ & AIoU $\uparrow$ & LPIPS $\downarrow$ & LDTW $\downarrow$ & DTW $\downarrow$ \\ \midrule GT strokes + Conv-Ptr-Net & -- & 73.62 & -- & -- & -- & -- \\ GT strokes + Transformer Pointer & -- & 92.27 & -- & -- & -- & -- \\ \midrule YOLO26 + Transformer Pointer & 91.63 & 52.75 & 0.7697 & 0.0228 & 2.74 & 250.27 \\ Mask2Former + Transformer Pointer & 94.64 & 64.25 & 0.7941 & 0.0152 & 2.05 & 180.94 \\ MaskDINO + Transformer Pointer & \textbf{94.80} & 65.90 & 0.7967 & 0.0138 & 1.93 & 155.80 \\ Ours (joint autoregressive) & 93.21 & \textbf{67.03} & \textbf{0.7970} & \textbf{0.0132} & \textbf{1.74} & \textbf{154.65} \\ \bottomrule \end{tabular} } \end{table*}

As shown in Table~\ref{tab:stroke_extraction_order}, the oracle
experiment confirms that post-hoc ordering itself can be highly
effective when correct stroke instances are available. However, once
the input strokes are produced by actual extraction models, missed,
redundant, split, or merged instances propagate directly to the
subsequent ordering stage. More importantly, higher stroke-extraction
accuracy does not necessarily translate into better stroke-order
recovery: although MaskDINO and Mask2Former achieve higher Mask
$AP_{50}$, our joint autoregressive formulation obtains the highest
strict character-level Order Acc. This distinction is also reflected
in downstream complete trajectory recovery, where our complete
pipeline achieves the best overall performance among the compared
systems. At the stroke-order level, these results support modeling stroke
identity and temporal order jointly during instance formation rather
than treating stroke ordering as an independent operation after
unordered extraction.

\begin{figure*}[t]
    \centering
    \includegraphics[
        width=0.98\textwidth,
        trim=20pt 20pt 20pt 20pt,
        clip
    ]{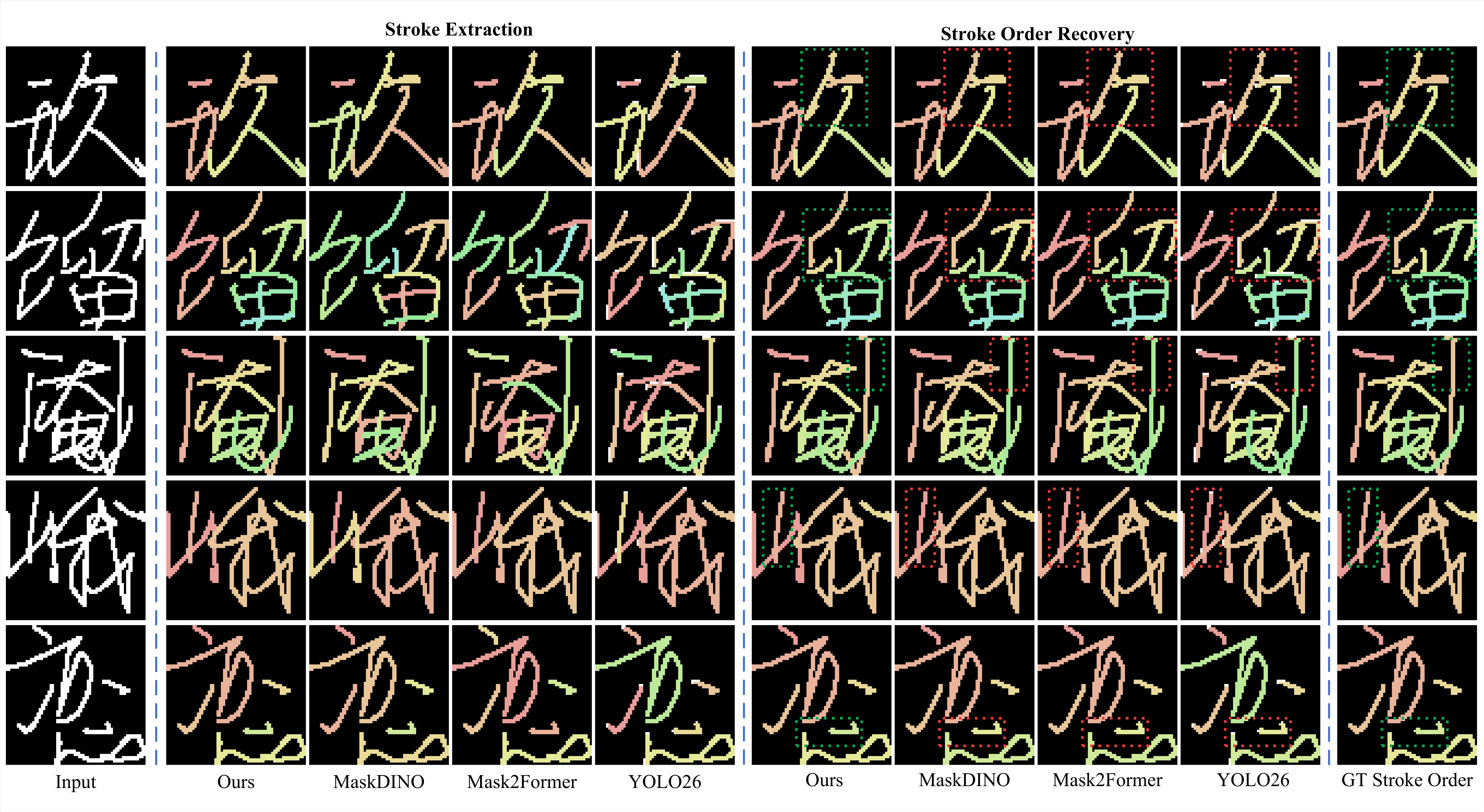}
    \caption{
        Qualitative comparison of stroke extraction and stroke-order
        recovery. The left group compares spatial stroke-instance predictions from
different models, while the right group compares the corresponding
recovered writing orders. Colors in the
        extraction results distinguish stroke instances only, whereas
        colors in the order-recovery results are indexed consistently
        according to writing order. Green and red dashed boxes
        highlight representative correct and incorrect local
        stroke-order recovery, respectively.
    }
    \label{fig:stagewise_comparison}
\end{figure*}

Figure~\ref{fig:stagewise_comparison} qualitatively illustrates the
same distinction. Similar spatial stroke extraction does not
necessarily lead to correct writing order, whereas our method more
consistently preserves the temporal organization required for
recovering the writing process.

\subsection{Complete Character Trajectory Recovery}
\label{subsec:complete_trajectory}

\begin{table}[t]
    \centering
    \caption{ Comparison with previously reported results on Chinese complete-character trajectory recovery. Baseline values follow the trajectory sampling and preprocessing settings of their original studies. Both of our settings use the complete two-stage recovery pipeline; full-point retains all original trajectory samples, while RDP $=3$ evaluates the influence of trajectory simplification. ``--'' denotes an unreported result. }
    \label{tab:complete_trajectory}
    \small
    \begin{tabular}{lcccc}
        \toprule
        Method
        & AIoU $\uparrow$
        & LPIPS $\downarrow$
        & LDTW $\downarrow$
        & DTW $\downarrow$ \\
        \midrule

        Cross-VAE
        & 0.146 & 0.402 & 13.64 & 1038 \\

        Kanji-Net
        & 0.326 & 0.186 & 5.51 & 443 \\

        DED-Net
        & 0.397 & 0.136 & 4.08 & 303 \\

        PEN-Net
        & 0.450 & 0.113 & 3.11 & 234 \\

        FINet
        & -- & -- & 1.78 & 225 \\

        Trajectory Transformer
        & 0.641 & 0.033 & 2.33 & 160 \\

        \midrule

        \textbf{Ours (full-point)}
        & 0.797
        & 0.013
        & 1.74
        & 155 \\

        Ours (RDP $=3$)
        & 0.773
        & 0.012
        & 1.52
        & 89.88 \\

        \bottomrule
    \end{tabular}
\end{table}

Table~\ref{tab:complete_trajectory} compares our method with previously
reported results on Chinese complete-character trajectory recovery.
The baseline values follow the trajectory sampling and preprocessing
settings adopted in their original studies, whereas our main result
retains all trajectory samples from the original online annotations.
Even without trajectory simplification, the proposed full-point model
achieves numerically better results than those reported by all compared
baselines across all four evaluation metrics, showing strong
complete-trajectory recovery performance under the full-point setting.

When the complete pipeline is trained and evaluated with RDP $=3$,
the sequence-alignment results are further improved. This observation
indicates that trajectory sampling density itself can influence the
measured recovery performance. We therefore further isolate this factor
from stroke extraction and stroke-order recovery errors in
Section~\ref{subsec:point_simplification}.

\subsubsection{Effect of Trajectory Sampling Density}
\label{subsec:point_simplification}

\begin{table}[t]
    \centering
    \caption{
        Effect of trajectory sampling density using ground-truth
        strokes. Recovered stroke trajectories are concatenated
        according to the ground-truth writing order and evaluated
        at the character level.
    }
    \label{tab:rdp_analysis}
    \small
    \setlength{\tabcolsep}{3.2pt}
    \resizebox{0.98\columnwidth}{!}{
    \begin{tabular}{lccccc}
        \toprule
        Trajectory setting
        & Point retention (\%)
        & AIoU $\uparrow$
        & LPIPS $\downarrow$
        & LDTW $\downarrow$
        & DTW $\downarrow$ \\
        \midrule

        Full-point
        & 100.00 & 0.7735 & 0.0191 & 1.622 & 102.97 \\

        RDP $=1$
        & 86.57 & 0.7773 & 0.0178 & 1.495 & 81.57 \\

        RDP $=2$
        & 78.11 & 0.7811 & 0.0168 & 1.343 & 66.07 \\

        RDP $=3$
        & 71.44 & 0.7851 & 0.0158 & 1.247 & 56.39 \\

        RDP $=4$
        & 63.55 & 0.7868 & 0.0149 & 1.118 & 44.47 \\

        RDP $=5$
        & 57.31 & 0.7891 & 0.0142 & 1.033 & 36.53 \\

        \bottomrule
    \end{tabular}
    }
\end{table}

To isolate the influence of trajectory sampling density from first-stage errors, Table~\ref{tab:rdp_analysis} evaluates within-stroke trajectory recovery using ground-truth strokes and reassembles the recovered strokes according to the ground-truth writing order before character-level evaluation. As trajectory simplification becomes stronger, all four evaluation metrics improve consistently, confirming that sampling density has a substantial influence on the measured recovery performance. 

One possible explanation for this trend lies in both sequence length
and target structure. Stronger RDP simplification removes more
redundant intermediate samples, thereby shortening the trajectory
sequence and reducing the number of points to be predicted. Meanwhile,
the retained samples are more concentrated at geometrically
informative locations that preserve the main trajectory shape, such as
endpoints and local turning regions. These changes may reduce the
complexity of the trajectory target and make trajectory generation and
alignment easier.

These results show that trajectory simplification can produce more
favorable recovery scores. We therefore use full-point recovery as
the primary setting, since it retains the complete sampled writing
process, while the RDP experiments are used to quantify how trajectory
simplification changes the evaluation results.

\subsection{Start/End Point Supervision and Start-Point Initialization}
\label{subsec:start_point_ablation}

We investigate two complementary components of the proposed framework:
start/end point prediction as auxiliary structural supervision in the
first stage, and the transfer of the predicted start point to initialize
within-stroke trajectory generation in the second stage. For variants without cross-stage start-point initialization, the
writing origin is predicted by the second-stage trajectory generator
from the isolated stroke.

\begin{table*}[t]
    \centering
    \caption{
        Ablation study of start/end point supervision and cross-stage
        start-point initialization. Start/End Sup. denotes auxiliary
        prediction of stroke start and end points in the first stage,
        while Start Init. denotes initialization of second-stage
        trajectory generation using the start point predicted by the
        first stage.
    }
    \label{tab:start_point_ablation}
    \small
    \resizebox{0.98\textwidth}{!}{
        \begin{tabular}{@{}lccccccc@{}}
            \toprule
            Variant
            & Start/End Sup.
            & Start Init.
            & Mask AP $\uparrow$
            & AIoU $\uparrow$
            & LPIPS $\downarrow$
            & LDTW $\downarrow$
            & DTW $\downarrow$ \\
            \midrule

            Baseline
            & $\times$ & $\times$
            & 84.989
            & 0.795
            & 0.0143
            & 1.82
            & 172 \\

            + Structural Points
            & $\checkmark$ & $\times$
            & 85.079
            & 0.795
            & 0.0144
            & 1.80
            & 165 \\

            Full Model
            & $\checkmark$ & $\checkmark$
            & 85.079
            & \textbf{0.797}
            & \textbf{0.0132}
            & \textbf{1.74}
            & \textbf{155} \\

            \bottomrule
        \end{tabular}
    }
\end{table*}

Table~\ref{tab:start_point_ablation} shows that the two components
affect trajectory recovery in different ways. Introducing start/end
point supervision leaves the aggregate Mask AP essentially unchanged,
but improves the sequence-alignment results even though the predicted
start point is not provided to the second stage. This suggests that
the auxiliary endpoint targets introduce direction-related structural
constraints into the shared first-stage representation, producing
stroke predictions that are more favorable for subsequent trajectory
recovery although the difference is not captured by the overall mask
overlap metric. When the predicted start point is further transferred
to the second stage, trajectory recovery improves consistently without
a corresponding change in Mask AP, demonstrating the additional
benefit of explicitly conditioning trajectory generation on the
writing origin recovered in the first stage.

\begin{figure*}[t]
    \centering
    \includegraphics[
        width=0.98\textwidth,
        trim=30pt 30pt 30pt 30pt,
        clip
    ]{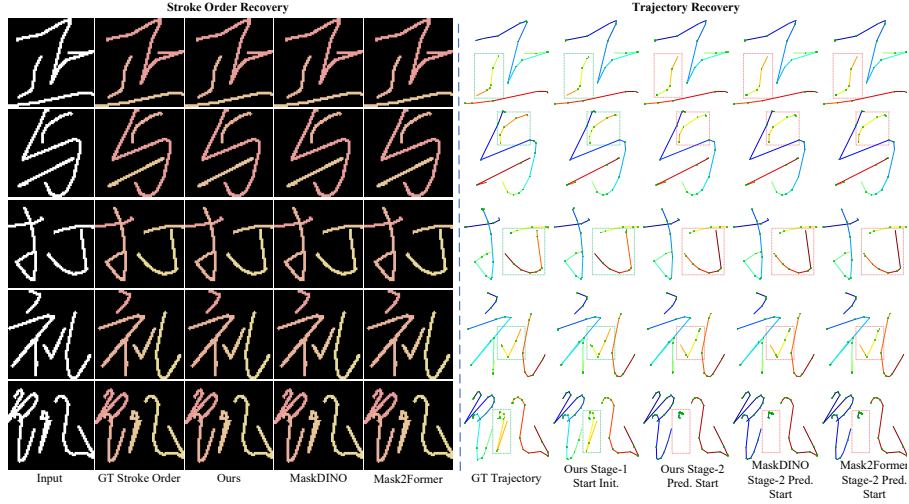}
    \caption{
        Qualitative comparison of different start-point conditions for
        trajectory recovery under correctly recovered stroke structures
        and writing orders. Only Ours with Stage-1 Start Init. uses the
        start point inferred by the first stage from the complete
        character and preceding writing state. For Ours with Stage-2
        Pred. Start and the MaskDINO- and Mask2Former-based pipelines,
        the writing origin is predicted by the same second-stage
        trajectory generator from the isolated stroke. Stroke colors
        encode writing order, while trajectory colors follow the
        cool-to-warm temporal convention introduced in
        Fig.~\ref{fig:writing_process}. Green and red dashed boxes
        highlight representative correct and incorrect writing-direction
        recovery, respectively.
    }
    \label{fig:start_point_comparison}
\end{figure*}

Figure~\ref{fig:start_point_comparison} further isolates the effect of
the start-point source. Even when stroke structure and inter-stroke
order are correctly recovered, trajectories whose writing origins are
predicted only from isolated strokes can still exhibit similar
directional errors. A static stroke mask specifies the spatial support
of a stroke but does not determine which endpoint was written first.
In contrast, the first-stage start point is inferred using the complete
character together with the preceding writing state, providing
directional information unavailable to the isolated second-stage
generator. Cross-stage start-point initialization therefore helps resolve
the writing-origin ambiguity before trajectory decoding: the first
stage determines \emph{where the stroke starts}, while the second stage
recovers \emph{how the pen subsequently moves}.

\subsection{Generalization to Unseen Chinese Character Categories}
\label{subsec:unseen_categories}

We further evaluate whether the learned writing-process representation
can generalize beyond the character categories observed during
training. The category-disjoint test set described in
Section~\ref{subsec:experimental_setup} contains only unseen Chinese
character categories, while the main validation set is reported for
reference.

\begin{table*}[t]
    \centering
    \caption{
        Generalization from seen to unseen Chinese character categories.
        Avg. strokes and Max. points/stroke characterize the structural
        complexity of the two evaluation sets.
    }
    \label{tab:unseen_categories}
    \small
    \resizebox{\textwidth}{!}{
        \begin{tabular}{lcccccccc}
            \toprule
            Evaluation set
            & \makecell{Avg.\\strokes}
            & \makecell{Max.\\points/stroke}
            & Mask AP $\uparrow$
            & \makecell{Order \\Acc. (\%) $\uparrow$}
            & AIoU $\uparrow$
            & LPIPS $\downarrow$
            & LDTW $\downarrow$
            & DTW $\downarrow$ \\
            \midrule

            Seen categories
            & 5.55 & 99
            & 85.079 & 67.03
            & 0.7970 & 0.0132
            & 1.74 & 155 \\

            Unseen categories
            & 7.32 & 134
            & 80.308 & 52.06
            & 0.7954 & 0.0153
            & 1.88 & 246 \\

            \bottomrule
        \end{tabular}
    }
\end{table*}

As shown in Table~\ref{tab:unseen_categories}, the unseen-category set
contains more strokes per character and longer within-stroke
trajectories, making both ordered stroke recovery and complete
trajectory reconstruction more challenging. Under this increased
structural complexity, the strict character-level Order Acc. decreases,
as any stroke extraction or ordering error invalidates the entire
character. Nevertheless, the recovered trajectories retain comparable spatial
coverage and perceptual consistency, while LDTW exhibits only a
moderate degradation. DTW degrades more noticeably on this structurally
more complex unseen-category set. These results suggest that the proposed model does not rely solely on
character-specific template memorization. Instead, the learned stroke-instance
representations, inter-stroke ordering relationships, and within-stroke
motion patterns can be recomposed to recover the writing process of
previously unseen Chinese character categories.

\subsection{English and Tamil Handwriting Trajectory Recovery}
\label{subsec:english_tamil}

We further evaluate the cross-language extensibility of the proposed
framework on English and Tamil handwriting. These two evaluations
represent different settings beyond the main Chinese experiments.
For English, the character categories are included during training,
while the test samples are written by different writers; the experiment
therefore evaluates whether the same writing-process recovery framework
remains effective on another language under the full-point trajectory
setting. For Tamil, both stages are fine-tuned on target-language
training data, providing a more challenging evaluation of adaptability
to a substantially different writing system.

The results of Cross-VAE, Kanji-Net, DED-Net, and PEN-Net are taken
from the comparisons reported in PEN-Net~\cite{chen2022pennet}, while
the results of Trajectory Transformer are taken from its corresponding
paper~\cite{lin2024trajectorytransformer}. These baseline values follow
the sampling and preprocessing settings of their respective studies
and are therefore treated as previously reported reference results.

\begin{table}[t]
    \centering
    \caption{
        Comparison with previously reported results on English
        handwriting trajectory recovery. Our method is evaluated under
        the full-point trajectory setting on test samples written by
        writers different from those used for training.
    }
    \label{tab:english_results}
    \small
    \begin{tabular}{lcccc}
        \toprule
        Method
        & AIoU $\uparrow$
        & LPIPS $\downarrow$
        & LDTW $\downarrow$
        & DTW $\downarrow$ \\
        \midrule

        Cross-VAE
        & 0.238 & 0.206 & 7.43 & 177 \\

        Kanji-Net
        & 0.356 & 0.121 & 5.98 & 150 \\

        DED-Net
        & 0.421 & 0.089 & 4.70 & 110 \\

        PEN-Net
        & 0.461 & 0.074 & 3.21 & 77 \\

        Trajectory Transformer
        & 0.608 & 0.035 & 3.06 & 74 \\

        \midrule

        Ours
        & \textbf{0.776}
        & \textbf{0.015}
        & \textbf{1.82}
        & \textbf{53} \\

        \bottomrule
    \end{tabular}
\end{table}

As shown in Table~\ref{tab:english_results}, our method achieves
numerically better results than those reported by all compared
baselines while retaining all original trajectory samples. Although the English character categories
are observed during training, the experiment extends trajectory
recovery beyond the main Chinese setting and additionally evaluates
different test writers. The results show that the same formulation can also
recover the spatial and temporal organization of English handwriting
under the full-point recovery setting.

\begin{table}[t]
    \centering
    \caption{
        Comparison with previously reported results on Tamil
        handwriting trajectory recovery. Both stages of our method are
        fine-tuned using Tamil training data and follow the Tamil
        experimental settings adopted in PEN-Net and Trajectory
        Transformer.
    }
    \label{tab:tamil_results}
    \small
    \begin{tabular}{lcccc}
        \toprule
        Method
        & AIoU $\uparrow$
        & LPIPS $\downarrow$
        & LDTW $\downarrow$
        & DTW $\downarrow$ \\
        \midrule

        Cross-VAE
        & 0.235 & 0.228 & 4.89 & 347 \\

        Kanji-Net
        & 0.340 & 0.163 & 3.04 & 234 \\

        DED-Net
        & 0.519 & 0.084 & 2.00 & 130 \\

        PEN-Net
        & 0.546 & 0.074 & 1.62 & 105 \\

        Trajectory Transformer
        & 0.637 & 0.048 & 1.50 & 84 \\

        \midrule

        Ours
        & \textbf{0.756}
        & \textbf{0.018}
        & \textbf{1.43}
        & \textbf{81} \\

        \bottomrule
    \end{tabular}
\end{table}

Table~\ref{tab:tamil_results} further shows that, after target-language
fine-tuning, the same framework can be adapted to Tamil
handwriting with substantially different glyph structures and writing
patterns. Together, the English and Tamil experiments demonstrate
cross-language extensibility under two complementary settings:
evaluation on English handwriting whose character categories are
included during training but whose test writers are different, and
adaptation to a different writing system through target-language
fine-tuning. Combined with the unseen-category
Chinese results, these findings support that recovering the writing
process through stroke instances, inter-stroke order, and within-stroke
motion provides a reusable formulation for writing-process recovery beyond a single
language or character set. Since English and Tamil use different
trajectory sampling settings, their absolute DTW and LDTW values are
not directly compared across datasets.

\section{Conclusion}
\label{sec:conclusion}

This work approaches offline-to-online handwriting trajectory recovery
from the perspective of recovering the writing process itself. Rather
than directly generating a complete character trajectory, we follow
the intrinsic hierarchy of handwriting and decompose recovery into
autoregressive ordered stroke instance prediction and within-stroke
trajectory generation. The first stage reformulates query-based
unordered set prediction as autoregressive ordered prediction, allowing
stroke extraction and stroke-order recovery to be performed within the
same instance-formation process. Start and end points further provide
direction-related structural constraints for the first-stage
representation, while the predicted start point is transferred across
stages to determine the writing origin before subsequent pen-tip motion
is generated.

The experiments consistently support this two-stage formulation.
Modeling stroke order during instance formation is more effective than
post-hoc ordering of independently extracted strokes, while cross-stage
start-point initialization further improves complete trajectory
recovery by reducing within-stroke writing-direction ambiguity. Even without trajectory simplification, our full-point model achieves
numerically better results than those reported by all compared
baselines. The controlled sampling-density analysis further
demonstrates that trajectory sampling density has a substantial
influence on measured recovery performance. In addition,
experiments on unseen Chinese character categories and on English and
Tamil handwriting demonstrate category-level generalization and
cross-language extensibility under different evaluation settings.
Overall, progressively recovering stroke structure, inter-stroke order,
writing origin, and within-stroke pen-tip motion provides an effective
way to recover the writing process hidden behind static handwriting.

    \section*{CRediT authorship contribution statement}

En-Guang Wang: Conceptualization, Methodology, Software, Data curation,
Investigation, Formal analysis, Validation, Visualization,
Writing -- original draft;
Yan-Ming Zhang: Conceptualization, Methodology, Supervision,
Writing -- review \& editing;
Fei Yin: Project administration, Supervision;
Cheng-Lin Liu: Supervision, Writing -- review \& editing.

\section*{Declaration of competing interest}

The authors declare that they have no known competing financial
interests or personal relationships that could have appeared to
influence the work reported in this paper.

\section*{Data availability}

The datasets used in this study are available from their respective
original sources. The processed data and implementation code are
available from the first author upon reasonable request.

\section*{Declaration of generative AI and AI-assisted technologies
in the manuscript preparation process}

During the preparation of this work, the authors used ChatGPT (OpenAI)
to assist with language refinement and manuscript organization.
After using this tool, the authors reviewed and edited the content as
needed and take full responsibility for the content of the publication.
    
    \bibliographystyle{elsarticle-num}
    \bibliography{references}
    
    \end{document}